\documentclass{optica-article}

\journal{opticajournal} 

\articletype{Research Article}

\usepackage{lineno}

\usepackage{booktabs}
\usepackage{multirow}
\usepackage[capitalize]{cleveref}
\usepackage{siunitx}
\usepackage{makecell}

\begin{document}

\title{Quantized neural network for complex hologram generation}

\author{Yutaka Endo,\authormark{1,*} Minoru Oikawa,\authormark{2} Timothy D. Wilkinson,\authormark{3} Tomoyoshi Shimobaba,\authormark{4} and Tomoyoshi Ito\authormark{4}}

\address{\authormark{1}Institute of Science and Engineering, Kanazawa University, Kakuma-machi, Kanazawa, Ishikawa 920-1192, Japan\\
\authormark{2}Science and Technology Unit, Natural Sciences Cluster, Kochi University, 2-5-1 Akebono-cho, Kochi 780-8520, Japan\\
\authormark{3}Electrical Engineering Division, Department of Engineering, University of Cambridge, 9 JJ Thomson Avenue, Cambridge CB3 0FA, UK\\
\authormark{4}Graduate School of Engineering, Chiba University, 1-33 Yayoi, Inage, Chiba 263-8522, Japan}

\email{\authormark{*}endo@se.kanazawa-u.ac.jp} 


\begin{abstract*}
Computer-generated holography (CGH) is a promising technology for augmented reality displays, such as head-mounted or head-up displays.
However, its high computational demand makes it impractical for implementation.
Recent efforts to integrate neural networks into CGH have successfully accelerated computing speed, demonstrating the potential to overcome the trade-off between computational cost and image quality.
Nevertheless, deploying neural network-based CGH algorithms on computationally limited embedded systems requires more efficient models with lower computational cost, memory footprint, and power consumption.
In this study, we developed a lightweight model for complex hologram generation by introducing neural network quantization.
Specifically, we built a model based on tensor holography and quantized it from 32-bit floating-point precision (FP32) to 8-bit integer precision (INT8).
Our performance evaluation shows that the proposed INT8 model achieves hologram quality comparable to that of the FP32 model while reducing the model size by approximately 70\% and increasing the speed fourfold.
Additionally, we implemented the INT8 model on a system-on-module to demonstrate its deployability on embedded platforms and high power efficiency.
\end{abstract*}

\section{Introduction}

Computer-generated holography (CGH) is a technique that controls light waves using computationally rendered holograms.
CGH has diverse applications in three-dimensional (3D) displays, bioimaging, and laser processing. 
One particularly promising application is augmented reality (AR) displays, including head-mounted or head-up displays \cite{chang2020_nextgeneration}.
Several studies have investigated CGH-based AR displays due to their potential to realize natural depth cues and simple optics \cite{wakunami2016_projectiontype,maimone2017_holographic,jang2018_holographic,sano2021_holographic,kim2022_holographic,aksit2023_holobeam}.

A major challenge in realizing practical CGH applications is the substantial computational demand required to achieve photorealistic images.
CGH requires numerical light propagation from a 3D scene to compute a hologram.
This calculation is highly non-local, meaning that a single geometric primitive affects many hologram pixels.
Therefore, the computational cost typically scales with the number of hologram pixels and the complexity of the 3D scene.
Additionally, photorealistic holographic images require physically accurate light propagation models and time-consuming iterative optimization, posing a trade-off between computational cost and image quality.
Considerable effort has been put into developing fast and high-quality CGH calculations, and recent progress is summarized in \cite{blinder2022_stateoftheart}.
Traditional CGH algorithms have been accelerated by approximating a light propagation model \cite{shimobaba2017_fast,shiomi2022_fast}, using efficient 3D scene representations \cite{shimobaba2009_simple,blinder2020_phase}, caching pre-computed results \cite{kim2008_effective,nishitsuji2015_fast}, and leveraging hardware accelerators \cite{sugie2018_highperformance,yamamoto2022_horn9,kim2019_singlechip,an2020_slimpanel,wang2020_hardware}.
Recently, the integration of neural networks for CGH has gained attention for its potential to overcome the trade-off between computational cost and image quality \cite{shimobaba2022_deeplearning,horisaki2018_deeplearninggenerated,horisaki2021_threedimensional,wu2021_highspeed,yang2022_diffractionengineered,shui2022_diffraction,liu2023_4kdmdnet}.
One seminal study is \textit{tensor holography}, which uses a convolutional neural network (CNN) to efficiently approximate point-based light propagation \cite{shi2021_realtime}.
Another notable study is \textit{HoloNet}, which uses a U-Net architecture and achieves high-fidelity CGH optimization in real time \cite{peng2020_neural,choi2021_neural}.
Although neural networks have enabled fast CGH algorithms, they currently rely on high-performance graphics processing units (GPUs) for real-time processing, leading to bulky systems and high power consumption.
Deploying them on computationally limited embedded systems, such as AR headsets and automotive head-up displays, requires more efficient models to reduce computational cost, memory footprint, and power consumption.

Neural network quantization is a promising approach for creating lightweight models suitable for computationally limited platforms \cite{jacob2018_quantization,wu2020_integer,gholami2022_survey}.
Quantization reduces the bit precision of weights and activations in a model to lower-precision formats, typically from 32-bit floating-point precision (FP32) to 8-bit integer precision (INT8).
This technique contributes to the reduction in model size and memory footprint.
Furthermore, quantized models can leverage low-precision arithmetic, leading to faster computations.
Although quantization is effective, its application in CGH has not been well investigated.
Tensor holography uses quantization to create a lightweight model \cite{shi2021_realtime}, but the details of the quantization method are not provided because the main focus of this study is not on quantization.
Moreover, the hologram quality of the quantized model has not yet been evaluated.

This study introduces quantization to a neural network-based CGH algorithm to create a lightweight model.
Our model is based on tensor holography, which computes complex holograms from RGB-D images.
We quantized the FP32 model to an INT8 model and evaluated its hologram quality, model sizes, and computational speed.
Our evaluation of hologram quality revealed that applying INT8 \textit{static quantization} to the original tensor holography model failed to yield satisfactory hologram quality.
Therefore, our model introduced architectural refinements, enabling its INT8 version to achieve hologram quality comparable to that of the FP32 version.
Additionally, we implemented the INT8 model on a system-on-module (SoM), AMD Kria K26, to demonstrate its deployability in embedded systems.
Our SoM implementation achieved four times higher power efficiency than a GPU.

\section{Method}

\subsection{Complex hologram generation and tensor holography}

In CGH, a 2D hologram is computed to reproduce a desired intensity distribution by diffracting incident coherent light.
Hologram generation requires the light wave field created by a target scene on the hologram plane, known as a complex hologram.
One popular algorithm for computing complex holograms is the point-based method (PBM), where the target scene is represented as a set of points, and the contribution of each point to the hologram is calculated.
Although this method is flexible and easy to implement, its computational cost typically scales with the number of hologram pixels and the number of points.
Additionally, incorporating occlusion handling into the point-based method, known as the occlusion-aware point-based method (OA-PBM), introduces the additional cost, making its real-time processing challenging.

Shi \textit{et al}. proposed tensor holography to accelerate complex hologram generation from an RGB-D image using a CNN \cite{shi2021_realtime}.
The CNN consists of successive application of a set of 3×3 convolutions with ReLU activations, enabling to efficiently approximate the OA-PBM.
They built a large-scale hologram dataset, MIT-CGH-4K, consisting of RGB-D images and their corresponding complex holograms computed by the OA-PBM to train the CNN.
The trained CNN can compute complex holograms of similar quality to those calculated using the OA-PBM but significantly faster.

\subsection{Model architecture and training}
\label{ssec:model-architecture-and-training}

\begin{figure}[htbp]
\centering\includegraphics[width=\linewidth]{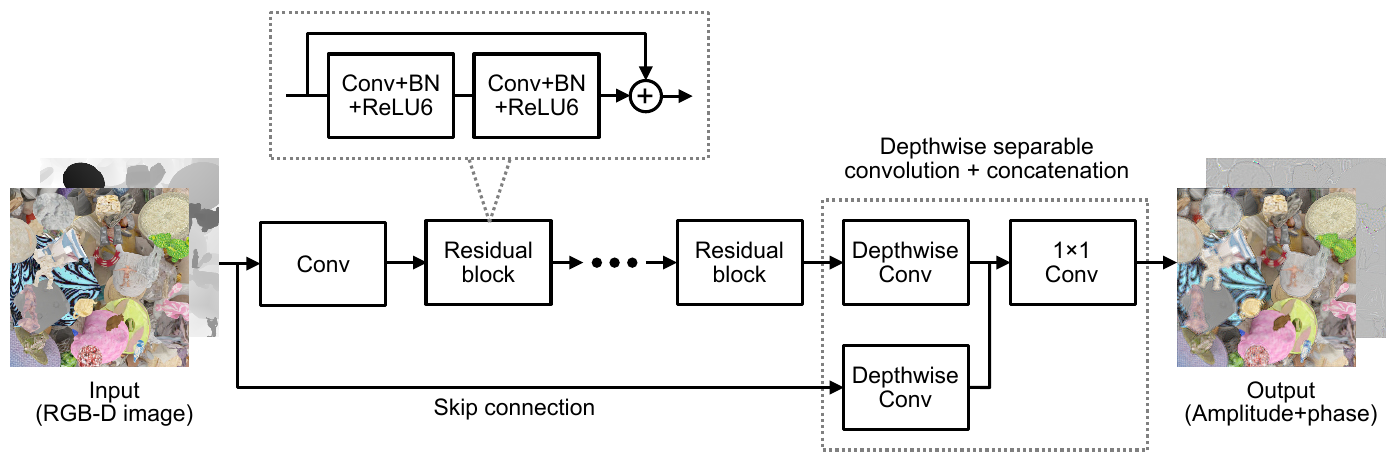}
\caption{Model architecture for computing complex holograms from RGB-D images. The residual blocks consist of two convolution layers with batch normalization (BN) and ReLU6 activation functions. The output of the final residual block and the skip-connected input are concatenated in the depthwise separable convolution block.}
\label{fig:model-arch}
\end{figure}

Building on tensor holography, we developed a refined CNN model to enhance its performance after quantization.
As illustrated in \cref{fig:model-arch}, our model consists of the stacks of residual blocks, each containing two 3×3 convolutions with batch normalization and ReLU6 activation functions.
ReLU6 is a rectified linear unit with the maximum threshold of six: $\operatorname{ReLU6}(x) = \min(\max(0, x), 6)$.
Our model concatenates the output of the final residual block with the skip-connected input in the depthwise separable convolution block, which consists of 3×3 depthwise and 1×1 pointwise convolutions \cite{howard2017_mobilenets}.
The Hardtanh activation function restricts the network output within the range $[0, 1]$.
Consistent with tensor holography, we used 29 convolution layers of 24 kernels each, except for the depthwise separable convolution block.

\begin{figure}[htbp]
\centering\includegraphics[width=\linewidth]{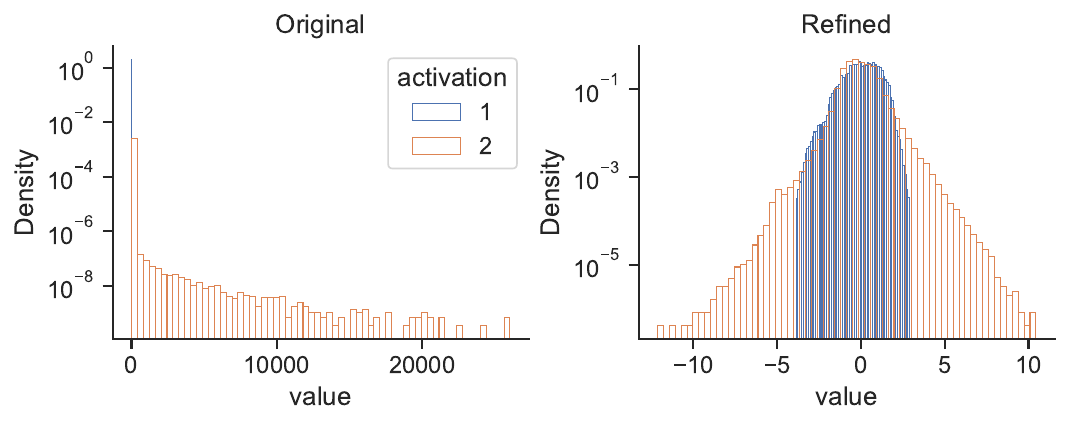}
\caption{Histograms of two activations that are concatenated in both the original and refined models.}
\label{fig:cat-hist}
\end{figure}

Our model includes three refinements over the original model used in tensor holography.
First, we used Hardtanh rather than Tanh because it is computationally less expensive and can be removed by appropriately setting the clipping range.
Second, we added a depthwise convolution with batch normalization before concatenating the output of the residual block and the skip-connected input.
This helps align the histograms of the two concatenated activations.
Finally, we used ReLU6 rather than ReLU to suppress the growth of the residual block outputs, thereby mitigating the accuracy degradation after quantization.
\Cref{fig:cat-hist} shows the histograms of two activations that are concatenated in both the original and refined models.
This example clearly demonstrates that the original model introduces an imbalance in the histograms between the skip-connected input and the output of the final residual block.
In contrast, our refined model addresses this issue by aligning the histograms, making them more consistent.

We implemented our model in PyTorch 2.1.0 and trained it using the MIT-CGH-4K dataset.
The image resolution and pixel pitch of the dataset were 384 × 384 pixels and \SI{8.0}{\um}, respectively.
The RGB channels correspond to wavelengths of 638, 520, and 450 nm.
We used 3,800 samples for training, 100 samples for validation, and 100 samples for testing.
The loss function is the sum of the amplitude and phase mean squared errors (MSEs):
$$
l = \sum_{n=1}^N \left[ \operatorname{MSE}(a_n, \hat{a}_n)
+ \frac{1}{2\pi} \operatorname{MSE}(\phi_n, \hat{\phi}_n) \right],
$$
where $a_n$ and $\phi_n$ are the $n$th target amplitude and phase, $\hat{a}_n$ and $\hat{\phi}_n$ are the $n$th predicted amplitude and phases, and $\operatorname{MSE}()$ calculates the MSE.
This loss function differs from the phase-corrected $\ell_2$ loss \cite{shi2021_realtime}, enabling us to evaluate the quality of both the amplitude and phase images without considering the global phase offsets.
Training was conducted for 100 epochs using the Adam optimizer with a learning rate of $1 \times 10^{-4}$ and a batch size of 2.

\subsection{Neural network quantization}

Quantization in a neural network is a technique that reduces the bit precision of weights and activations.
Specifically, quantization converts FP32, used in most machine learning frameworks for model parameters, to lower-precision formats such as 16-bit floating-point precision (FP16), 8-bit or 4-bit integers precision (INT8 and INT4), and ternary or binary values.
Reducing the bit widths benefits the model size and memory footprint.
Furthermore, computational speed increases if the processor supports high-throughput math pipelines in low-bit formats.
However, reducing bit precision may reduce accuracy, particularly for small models.
Nonetheless, most neural network models are over-parameterized, meaning that appropriate quantization can reduce the bit widths without significant accuracy degradation.

We begin by introducing quantization and dequantization operations on model weights and activations.
Here, we focus on uniform integer quantization, where a floating point number $x$ is mapped to a $b$-bit signed integer $x_q \in [-2^{b-1}, 2^{b-1}-1]$.
The quantization function is defined as:
$$
Q(x) = \operatorname{round}\left(\frac{x}{S}\right) + Z,
$$
where $\operatorname{round}()$ rounds to the nearest integer, and $S$ and $Z$ are quantization parameters called the scaling factor and zero-point, respectively.
The reconversion from an integer to a floating-point number uses a dequantization function defined as:
$$
\tilde{Q}(x_q) = (x_q - Z) \cdot S,
$$
where $S$ is determined by the ratio of the input and output ranges as
$$
S = \frac{\beta - \alpha}{2^b - 1},
$$
where $[\alpha, \beta]$ is the clipping range of the input (i.e., the boundaries of the permissible inputs).
The process of selecting a clipping range is often referred to as \textit{calibration}.
$Z$ is chosen to be zero for \textit{symmetric quantization} and nonzero for \textit{asymmetric quantization}.
Asymmetric quantization uses
$$
Z = -\operatorname{round}\left(\frac{\alpha}{S}\right) - 2^{b-1}.
$$
Quantized weights and activations enable multiply-accumulate (MAC) operations with low precision, which is computationally intensive in most neural networks.

The clipping ranges of weights can be computed statically, but those of activations vary for each input and can be determined either dynamically or statically.
\textit{Dynamic quantization} calculates the range for each activation during inference and quantizes it on the fly.
By contrast, \textit{static quantization} calculates the range using a representative dataset (i.e., calibration dataset) before inference to pre-quantize the activations.
While dynamic quantization can achieve higher accuracy than static quantization, it adds computational overhead for calculating the clipping range.
Additionally, numerous model deployment tools for embedded platforms only support statically quantized models and do not accommodate dynamically quantized models.

The straightforward quantization workflow is \textit{post-training quantization} (PTQ), which trains an FP32 model and then determines the quantization parameters of the pre-trained FP32 model without additional fine-tuning.
While PTQ is simple and easy to apply, it may significantly degrade accuracy depending on the parameter distribution.
By contrast, \textit{quantization-aware training} (QAT) retrains the pre-trained FP32 model while accounting for the quantization of the weights and activations.
QAT often achieves higher accuracy than PTQ, but it requires additional training time.

We quantized our FP32 model using PyTorch FX Graph Mode Quantization \cite{pytorch_quant}.
The quantization methods were \textit{post-training dynamic quantization} (PTDQ), \textit{post-training static quantization} (PTSQ), and static QAT, which are supported by the quantization framework.
PTDQ and PTSQ are PTQ methods using dynamic and static quantization, respectively.
The target backend was QNNPACK, and its default settings were used for calibration.
The calibration dataset for PTSQ consisted of 100 samples from the training dataset.
For QAT, we further trained the pre-trained FP32 models for 20 epochs with a reduced learning rate of $1 \times 10^{-6}$ and statically quantized the weights and activations.

\section{Results}

We evaluated the performance of our FP32 and quantized INT8 models in terms of output quality, model size, and speed. For comparison, we also assessed the original tensor holography model, which was trained and quantized under the same conditions as our model. In addition, we implemented our statically quantized INT8 model on a SoM platform and evaluated its speed and power consumption.

\subsection{Hologram quality}

\begin{figure}[htbp]
\centering\includegraphics[width=0.8\linewidth]{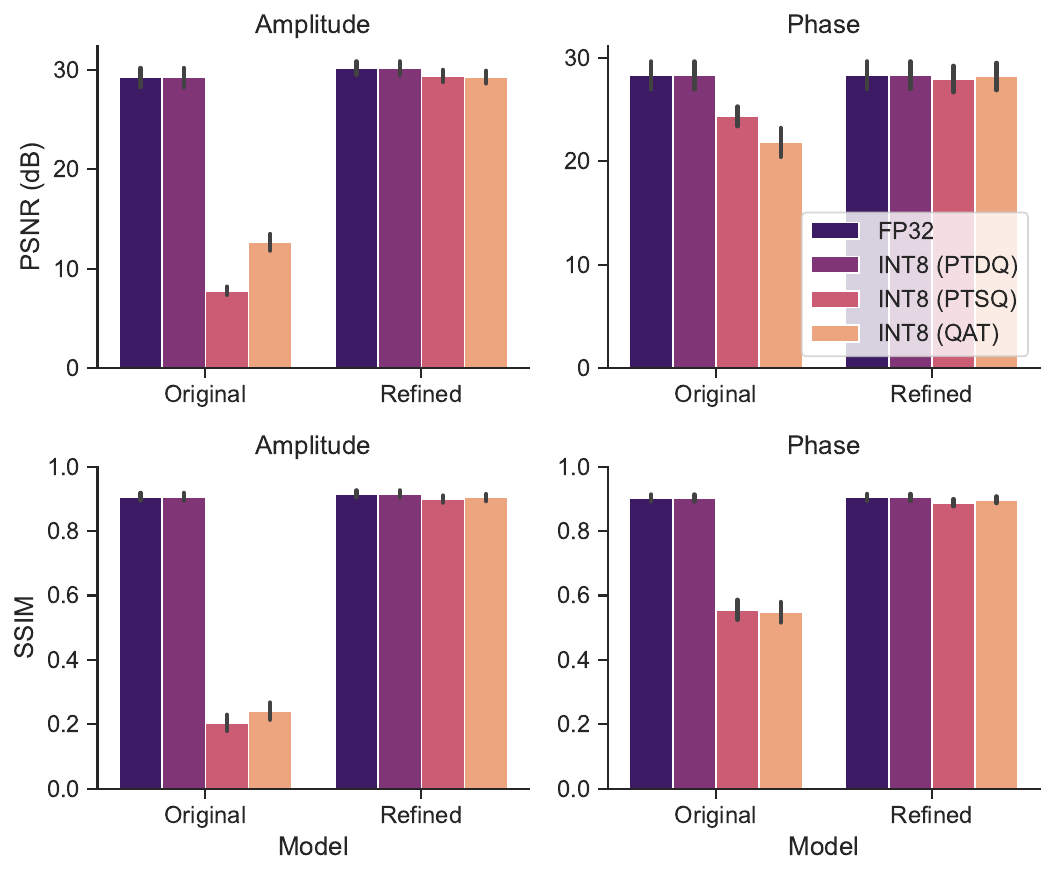}
\caption{Hologram quality of the original tensor holography model and our refined model in FP32 and INT8 with PTDQ, PTSQ, and QAT. PSNR and SSIM are averaged over the test dataset.}
\label{fig:hologram-quality}
\end{figure}

We evaluated the image quality of output holograms from the FP32 and INT8 models.
The image quality was assessed using the peak signal-to-noise ratio (PSNR) and the structural similarity index measure (SSIM).
These metrics were applied to both the amplitude and phase images of the complex hologram.
\Cref{fig:hologram-quality} shows the average hologram quality for each model over the test dataset.
In FP32 precision, both the original tensor holography model and our refined model had similar quality.
In INT8 precision, both the original and refined PTDQ models produced holograms nearly identical to those of the FP32 models.
However, the original PTSQ model showed a significant decrease in hologram quality compared to FP32.
Moreover, QAT did not improve this PTSQ model, suggesting the original model has architectural limitations that result in a significant drop in accuracy after static quantization.
In contrast, all refined INT8 models achieved comparable hologram quality to that of FP32.
The improvement with QAT was minimal since the refined PTSQ model already exhibited sufficient hologram quality.
As shown in \cref{ssec:model-architecture-and-training}, the significant accuracy drop in the original model is mainly caused by an imbalance in the histograms between the skip-connected input and the output of the final residual block.
Our refined model addresses this issue by using ReLU6 to suppress the growth of the residual block outputs and by aligning the histograms through the convolution layers prior to concatenation.
These results demonstrate that the decrease in hologram quality caused by INT8 static quantization can be minimized through our architectural enhancements.


\subsection{Reconstructed images}

\begin{figure}[htbp]
\centering\includegraphics[width=\linewidth]{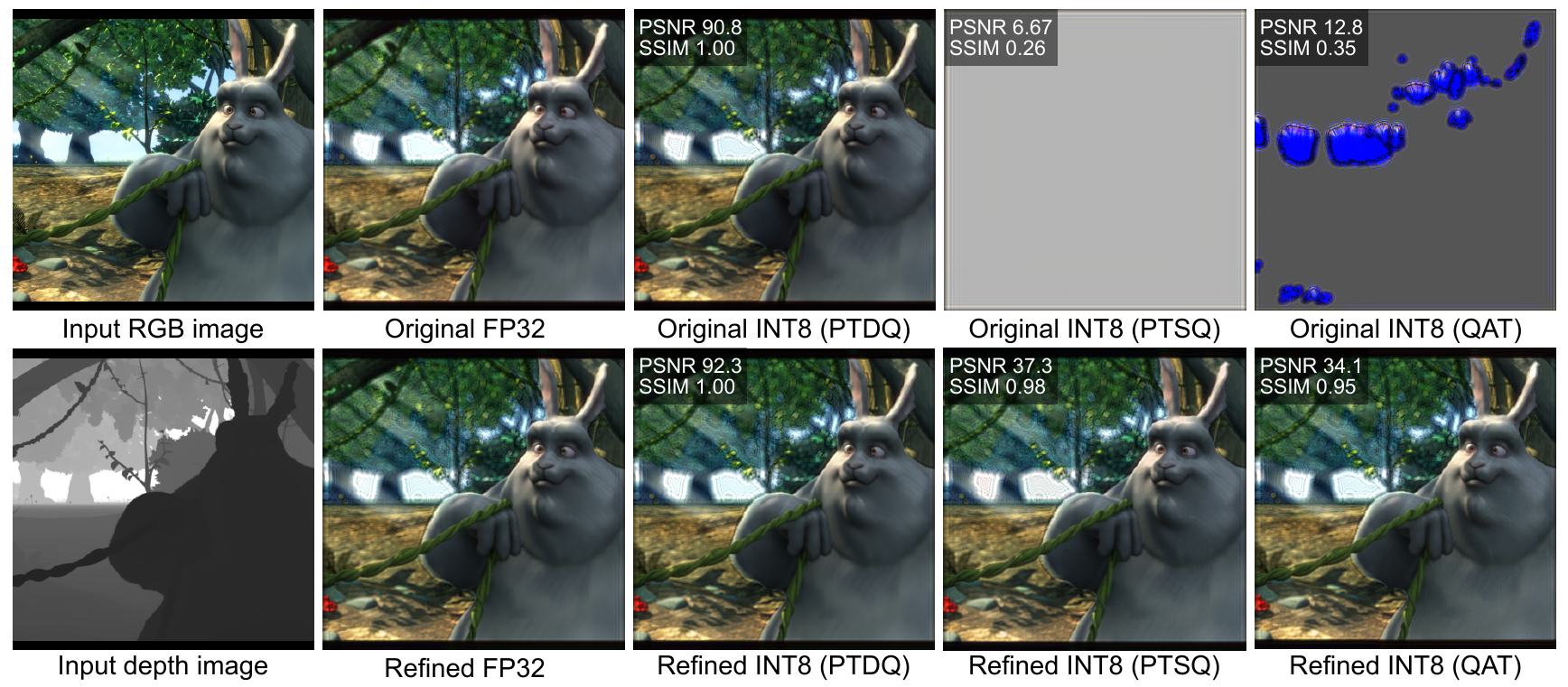}
\caption{Reconstructed images from complex holograms computed by original tensor holography model and our refined model in FP32 and INT8 with PTDQ, PTSQ, and QAT. The PSNR and SSIM are computed between the reconstructed images from the INT8 and FP32 models. \copyright 2008, Blender Foundation / www.bigbuckbunny.org.}
\label{fig:recon-images}
\end{figure}

We reconstructed images using the complex holograms output from the FP32 and INT8 models in simulation.
\Cref{fig:recon-images} shows the reconstructed images from the output complex holograms.
The resolution of the input RGB-D data was 384 × 384 pixels.
The angular spectrum method was used to compute the reconstructed images 2 mm away from the hologram plane.
We computed the PSNR and SSIM between the reconstructed images from the INT8 and FP32 models to evaluate the impact of quantization.
The results show that the original INT8 model with PTDQ accurately reconstructs the images, while those with PTSQ and QAT completely failed to do so.
However, even after static quantization, our refined INT8 models achieved comparable image quality to the non-quantized FP32 model.

\subsection{Model size, latency, and frame rate}

We evaluated the model size, latency, and frame rate of the FP32 and INT8 models, as shown in \cref{tbl:model-performance}.
In this evaluation, the INT8 models were statically quantized, because they are faster than their dynamic quantized counterparts.
The model sizes are represented by the file size of \texttt{state\_dict} of each model, which is a Python dictionary object that contains the model parameters used in PyTorch.
Latency refers to the computation time required to process a single image frame, whereas frame rate represents the number of image frames processed per second, expressed as frames per second (fps).
For the latency evaluation, we set the batch size to 1.
For the frame rate evaluation, we selected the best-performing batch size from 1, 2, 4, 8, 16, and 32.
The resolution of the input RGB-D images was set to 1280 × 720 pixels, corresponding to high-definition SLMs.
We executed the models using PyTorch with the QNNPACK backend.
Since our implementation did not support quantized inference on GPUs, we ran the INT8 model inference on an AMD Ryzen 7 5800 processor.
The results show that both the original and refined models had almost the same model size, latency, and frame rate.
The slight decrease in latency and frame rate appears to be attributed to the overhead of the depthwise and pointwise convolutions.
Compared to the FP32 models, both INT8 models demonstrated an approximately 70\% reduction in size and four times faster speed.
Thus, we demonstrated that quantization is an effective method for reducing the model size and computation time.

\begin{table}[htbp]
\caption{Model size, latency, and frame rate of the original tensor holography model and our refined model in FP32 and INT8 with static quantization. The input image resolution is 1280 × 720 pixels.}
\label{tbl:model-performance}
\centering
\begin{tabular}{ccccc}
\toprule
Model & Precision & Size (kB) & Latency (ms) & Frame rate (fps) \\
\midrule
Original & FP32 & 629 & 1,696 & 0.633 \\
Original & INT8 & 194 & 396  & 2.537 \\
\midrule
Refined  & FP32 & 631 & 1,714 & 0.607 \\
Refined  & INT8 & 198 & 427  & 2.360 \\
\bottomrule
\end{tabular}
\end{table}

\subsection{Deployment on embedded platforms}

To demonstrate the feasibility of deploying hologram generation models on embedded platforms, we implemented our refined INT8 model using the AMD Kria KV260 Vision AI Starter Kit and evaluated its performance.
The KV260 is a development platform for the Kria K26 SoM, which features an AMD Zynq Ultrascale+ MPSoC, integrating an FPGA fabric and ARM CPU cores.
To implement our model on the KV260 board, we used Vitis AI 3.5, a toolchain that enables the implementation of neural network models on AMD platforms.
In Vitis AI, neural networks run on a deep learning processing unit (DPU), a highly configurable and scalable accelerator for deep learning inference that can be implemented in an AMD programmable logic fabric.
Vitis AI provides DPU IPs and tools for model deployment on AMD platforms, including a quantizer, optimizer, and compiler.
We quantized our FP32 model using the Vitis AI Quantizer, compiled the quantized model using the Vitis AI Compiler, and ran it on a DPUCZDX8G B3136 DPU implemented on the KV260 board.

\Cref{tbl:device-performance} shows the latency, frame rate, and power consumption on the KV260 and a PC with an AMD Ryzen 7 5800 processor and an NVIDIA RTX 3090 for comparison.
The resolution of the input RGB-D images for the evaluation was set to 1,280 × 720 pixels.
For the frame rate evaluation, we set the batch size to 1 for KV260, while for the Ryzen 7 5800 and RTX 3090, we selected the best-performing batch size from 1, 2, 4, 8, 16, and 32.
The power consumption of the KV260 board and PC was measured using the \texttt{xlnx\_platformstats} command and a TAP-TST8N watt monitor, respectively.
The power consumption of the PC in the idle state was approximately 70–80 W.
The results show that although the KV260 was slower than the Ryzen 7 5800 and RTX 3090, its power consumption was significantly lower.
The computation efficiency of the INT8 model on the KV260, measured in fps per watt, was four times higher than that of the FP32 model on the RTX 3090.
Thus, we successfully implemented our INT8 model on an embedded platform and demonstrated its potential to achieve efficient CGH calculations.

\begin{table}[htbp]
\caption{Latency, frame rate, and power consumption of our INT8 model on the SoM (KV260) and PC (Ryzen 7 5800 and RTX 3090). The input image resolution is 1,280 × 720 pixels.}
\label{tbl:device-performance}
\centering
\begin{tabular}{cccccc}
\toprule
Platform & Precision & \makecell{Latency \\ (ms)} & \makecell{Frame rate \\ (fps)} & \makecell{Power \\ (W)} & \makecell{Efficiency \\ (fps/W)} \\
\midrule
KV260        & INT8 & 559.7 & 1.787 & 8.058 & 0.2218 \\
Ryzen 7 5800 & INT8 & 396.0 & 2.537 & 160   & 0.0159 \\
RTX 3090     & FP32 & 40.0  & 25.35 & 470   & 0.0539 \\
\bottomrule
\end{tabular}
\end{table}

\section{Discussion}

Our performance evaluation demonstrates that quantization effectively reduces model size and accelerates execution speed in a hologram generation model.
A prior study that used quantization did not provide details of the quantization settings and evaluation \cite{shi2021_realtime}.
This study provides evidence of the utility of quantization in CGH, contributing to the implementation of neural network-based CGH algorithms on computationally limited platforms.
In the following, we list some limitations of this study.

\subsection{Training and quantization settings}

To simplify the experiments, our training settings differed slightly from those used in the original tensor holography \cite{shi2021_realtime}.
While the original work used a combination of the phase-corrected $\ell_2$ loss and focal stack loss as the loss function, we used the amplitude and phase MSEs to ease the evaluation of hologram quality and achieve deterministic results.
Additionally, we reduced the number of training epochs to 100 rather than 1,000, as in the original study, to rapidly test our models.
These modifications do not significantly affect the main findings of this study.
However, the SSIM and PSNR values may vary under different training settings.

While the tensor holography paper does not provide the details on its quantization settings \cite{shi2021_realtime}, this study likely used different quantization settings (i.e., PyTorch's default).
The choice of quantization method can significantly affect accuracy.
This study did not investigate the optimal quantization method, leaving room for achieving better accuracy using other quantization methods.
Furthermore, it would be worthwhile to investigate more aggressive quantization using lower-precision formats than INT8, such as INT4, ternary, and binary formats.

\subsection{Image quality evaluation}

Our evaluation of hologram quality used PSNR and SSIM metrics on the complex holograms.
Notably, these metrics do not represent the quality of the reconstructed images themselves.
The reconstructed images were evaluated qualitatively through simulations using complex holograms.
This allows us to specifically evaluate the impact of quantization on the reconstructed images, independent of factors such as hologram encoding and optical setups.
Therefore, the reconstructed images in a real setup would likely be worse because complex holograms must be encoded into either amplitude-only or phase-only holograms, and the real setup has optical aberrations.
In future studies, we aim to evaluate the reconstructed images using a real setup.

\subsection{Model implementation}

Our implementation of the quantized models was not optimized and did not support GPU backends. 
Using other execution environments, such as ONNX Runtime and TensorRT, can improve the inference speed and enable quantized models to run on GPUs.

Although the SoM implementation using Vitis AI and DPU demonstrates its power efficiency, the frame rate is currently insufficient for real-time hologram generation.
In addition, it faces limitations in accepting input images of 1,920 × 1,080 pixels due to memory constraints.
To achieve real-time and high-quality hologram generation on embedded platforms, faster models with lower memory footprints must be developed.
The SoM implementation has also room for performance optimization.
For example, exploring DPU architectures other than the DPUCZDX8G B3136 and increasing the number of DPU cores could potentially improve the inference speed.

\section{Conclusion}

In this study, we demonstrated the INT8 quantization of a tensor holography-based model to reduce both the model size and computation time.
As INT8 static quantization of the original model causes a significant decrease in hologram quality, we introduced architectural refinements.
Our refined INT8 model achieved holography quality comparable to that of FP32.
The performance evaluation demonstrated that quantization effectively reduced the model size and execution time.
Additionally, we implemented our INT8 model on the Kria K26 using Vitis AI, demonstrating its deployability on an embedded platform and highlighting its high power efficiency.
We believe that this study provides essential insights for deploying neural network-based CGH algorithms on embedded platforms, leading to CGH-based AR displays with comfortable 3D views and compact form factors.

\begin{backmatter}

\bmsection{Funding} Japan Society for the Promotion of Science (JP22K17908, JP22KK0183).

\bmsection{Acknowledgments} Portions of this work were presented at the Computational Optical Sensing and Imaging in 2024, CTh4B.6.

\bmsection{Disclosures} The authors declare no conflicts of interest.

\bmsection{Data availability} Data underlying the results presented in this paper are not publicly available at this time but may be obtained from the authors upon reasonable request.

\end{backmatter}


\bibliography{references}






\end{document}